\def\year{2021}\relax
\documentclass[letterpaper]{article} 
\usepackage{aaai21}  
\usepackage{times}  
\usepackage{helvet} 
\usepackage{courier}  
\usepackage[hyphens]{url}  
\usepackage{graphicx} 
\usepackage{natbib}  
\usepackage{caption} 
\usepackage[switch]{lineno}
\usepackage{xcolor}
\usepackage{amsmath}
\usepackage{multirow}
\usepackage{array}
\title{FCFR-Net: Feature Fusion based Coarse-to-Fine Residual Learning for \\ Depth Completion}

\author{
    Lina Liu$^{1,2}$,
    Xibin Song$^{2,3}$,
    Xiaoyang Lyu$^{1}$,
    Junwei Diao$^{1}$,
    Mengmeng Wang$^{1}$,
    Yong Liu$^{1}$\footnote{Corresponding author} and
    Liangjun Zhang$^{2,3}$
    \\
}
\affiliations{
    \textsuperscript{\rm 1}Institute of Cyber-Systems and Control, Zhejiang University, China \\ 
    \textsuperscript{\rm 2}Baidu Research, China \\
    \textsuperscript{\rm 3}National Engineering Laboratory of Deep Learning Technology and Application, China
    \\
    \{linaliu, shawlyu, djw, mengmengwang\}@zju.edu.cn, \{songxibin, liangjunzhang\}@baidu.com, yongliu@iipc.zju.edu.cn\\
    
}

\begin{document}

\maketitle
\renewcommand{\thefootnote}{\fnsymbol{footnote}} 

\begin{abstract}
Depth completion aims to recover a dense depth map from a sparse depth map with the corresponding color image as input. Recent approaches mainly formulate depth completion as a one-stage end-to-end learning task, which outputs dense depth maps directly. However, the feature extraction and supervision in one-stage frameworks are insufficient, limiting the performance of these approaches. To address this problem, we propose a novel end-to-end residual learning framework, which formulates the depth completion as a two-stage learning task, i.e., a sparse-to-coarse stage and a coarse-to-fine stage. First, a coarse dense depth map is obtained by a simple CNN framework. Then, a refined depth map is further obtained using a residual learning strategy in the coarse-to-fine stage with a coarse depth map and color image as input. Specially, in the coarse-to-fine stage, a channel shuffle extraction operation is utilized to extract more representative features from the color image and coarse depth map, and an energy based fusion operation is exploited to effectively fuse these features obtained by channel shuffle operation, thus leading to more accurate and refined depth maps. We achieve SoTA performance in RMSE on KITTI benchmark. Extensive experiments on other datasets future demonstrate the superiority of our approach over current state-of-the-art depth completion approaches.
\end{abstract}

\section{Introduction}

Depth is considered as one of the most fundamental information in many applications, including robotics \cite{liao2017parse}\cite{Song_2019_CVPR}, augmented reality \cite{dey2012tablet}\cite{Song_2020_CVPR}, virtual reality \cite{armbruster2008depth} and SLAM \cite{wang2016framework}. Various depth sensors such as 3D Lidar, depth cameras and stereo cameras have been developed to obtain depth information. For autonomous driving, 3D Lidar is commonly used because it can obtain accurate depth information in centimeter-level accuracy. However, due to the inherent characteristics of Lidar devices, the captured depth information is usually sparsely distributed, which largely limits the performances of Lidar-based applications.

\begin{figure}[t!]
	\centering
	\setlength{\tabcolsep}{1pt}
	\def\mywidth{0.16}
	\begin{tabular}{ccccc}
		\includegraphics[width=0.32\linewidth]{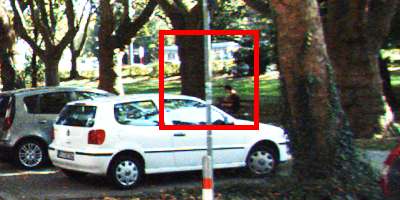} &
		\includegraphics[width=\mywidth\linewidth]{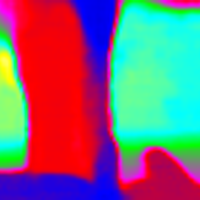} &
        \includegraphics[width=\mywidth\linewidth]{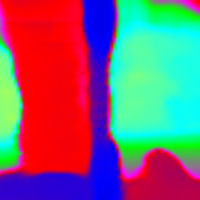} &
        \includegraphics[width=\mywidth\linewidth]{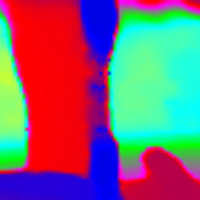} &
        \includegraphics[width=\mywidth\linewidth]{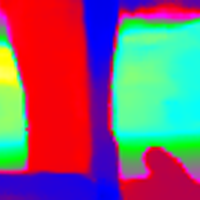} \\

        \includegraphics[width=0.32\linewidth]{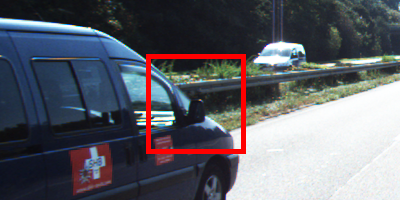} &
		\includegraphics[width=\mywidth\linewidth]{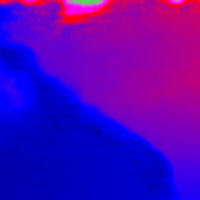} &
        \includegraphics[width=\mywidth\linewidth]{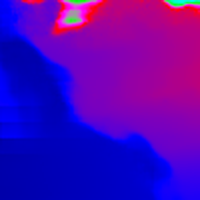} &
        \includegraphics[width=\mywidth\linewidth]{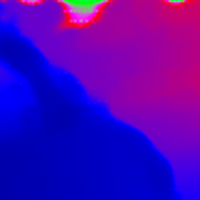} &
        \includegraphics[width=\mywidth\linewidth]{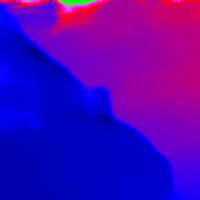} \\
        Color Image & (a) & (b) & (c) & (d) \\

	\end{tabular}

	\caption{Comparison with SoTA approaches. (a) STD~\cite{ma2019self}, (b) DeepLidar \cite{qiu2019deeplidar}, (c) CSPN++ \cite{cheng2020cspn++}, (d) Ours. }
	\label{fig:teaser_image}
\end{figure}

In order to obtain a dense and accurate depth map at a low cost, the task of depth completion draws more and more attention. Depth completion aims to recover a dense depth map from a sparse depth map obtained from Lidar or other depth sensors. Recently, various of effective depth completion approaches have been proposed, including sparse depth based approaches~\cite{uhrig2017sparsity}\cite{chodosh2018deep}\cite{Lu_2020_CVPR}\cite{Eldesokey_2020_CVPR} and image-guided based approaches~\cite{ma2019self}\cite{qiu2019deeplidar}\cite{Yang_2019_CVPR}\cite{Imran_2019_CVPR}\cite{Qu_2020_WACV}\cite{tang2019learning}\cite{cheng2020cspn++}\cite{li2020multi}\cite{liu2021learning}. However, due to lack of complementary cues of color information, the resulting depth maps of sparse depth based approaches are inevitably blurred with unclear boundaries. More recent image-guided based approaches explore color images to guide depth completion and various features fusion strategies have been proposed. 
However, image-guided based approaches mainly take depth completion as a one-stage task, and the feature extraction and information supervision are insufficient. Thus depth details are failed to be recovered. As illustrated in Fig.~\ref{fig:teaser_image} (a) to (c), the obtained depth maps either suffer from blurred edges or lose depth details.

To solve these problems, we propose an effective framework, named FCFR-Net, which tackles depth completion as a two-stage task, i.e., a sparse-to-coarse stage and a coarse-to-fine stage. The sparse-to-coarse stage first interpolates the sparse depth maps using simple CNN frameworks, and coarse dense depth maps can be obtained, which guarantees that more consecutive information can be provided in the next stage. Note that all the commonly used sparse-to-dense frameworks can be utilized in the sparse-to-coarse stage, and to reduce the complexity, we use the supervised network of STD~\cite{ma2019self} in our paper. Then the obtained coarse dense depth maps and corresponding color images are fed into the coarse-to-fine stage. To sufficiently fuse features extracted by color and depth information, a channel shuffle extraction operation and an energy based fusion operation are combined into a residual learning framework. The channel shuffle operation first interleaves color and depth features at multi-scale feature levels by mixing and disrupting the features of color and depth information at the channel level, and the energy based fusion operation further effectively fuses features obtained by channel shuffle operation. Hence, more representative features can be obtained, and more accurate depth completion results can be expected. The residual learning framework can further improve the performance of depth completion. As demonstrated in Fig.~\ref{fig:teaser_image} (d), compared with previous approaches, depth maps with sharper boundaries and more depth details can be obtained by our approach.

The main contributions of this paper can be summarized as:
\begin{itemize}
\item We formulate the problem of depth completion as a two-stage task, and a coarse-to-fine residual learning based framework is proposed, which contains a sparse-to-coarse stage and a coarse-to-fine stage. The sparse-to-coarse stage interpolates coarse dense depth maps, and the coarse-to-fine stage further refines the depth maps.
\item A channel shuffle extraction operation is proposed, which can effectively fuse the features of color and depth information at the multi-scale feature levels, and greatly improve the depth completion performance.
\item A energy based fusion operation is utilized to further sufficiently fuse the features obtained by channel shuffle extraction, thus achieves better performance.
\end{itemize}

We achieve SoTA performance in RMSE on KITTI benchmark, and results on NYUv2 dataset also demonstrates the superiority of our approach.

\section{Related Work}

\subsubsection{Sparse Depth based Approaches} Sparse depth can be used as input to get dense ones without image guidance~~\cite{uhrig2017sparsity}\cite{chodosh2018deep}, and most recent approaches~\cite{van2019sparse}\cite{li2020multi}\cite{qiu2019deeplidar}\cite{tang2019learning}\cite{park2020non} usually get dense depth map with sparse depth and image data as input. All of them encode the invalid values of sparse input with zeros. However, discontinuous values limit the performance of these approaches. Meanwhile, the sparse depth can also be interpolated along with the gravity~\cite{liao2017parse}\cite{chen2018estimating}, the invalid values are populated with non-zero values. Through these operations can effectively avoid the limitation in convolution learning, depth details and semantic information are lost. 

\subsubsection{Signal Level Fusion} \cite{ma2019self} use a ResNet~\cite{he2016deep} based autoencoder network to predict a dense depth map. The sparse depth map and image are directly connected as an input to the network at the signal level. To get a more accurate depth map, \cite{cheng2018depth}\cite{cheng2020cspn++} propose a novel convolutional spatial propagation network (CSPN) to learn the afﬁnity matrix for depth prediction. This work adopts a general CNN structure and adds post-processing to get a sharp result. All the above methods directly merge the image and sparse depth at the signal level, and then obtain more accurate results through post-processing.

\begin{figure*}[htb]
	\centering
	\includegraphics[width=0.9\linewidth]{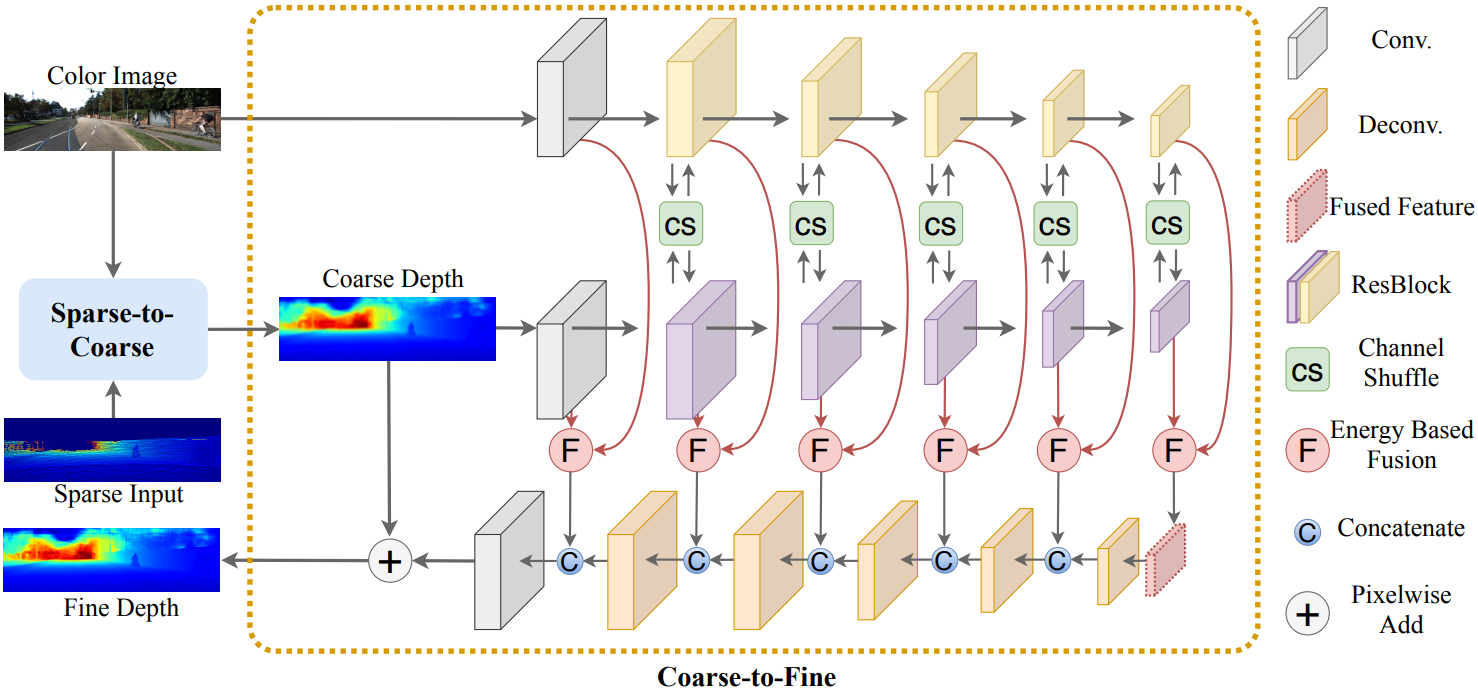}
	
	\caption{Overview of network architecture. The whole network architecture includes two parts: Sparse-to-Coarse stage (blue area) and Coarse-to-Fine stage (orange area). A simple CNN framework is used for the Sparse-to-Coarse stage. In the Coarse-to-Fine stage, color image and coarse depth are extracted by two encoder branches. Channel shuffle is utilized to mix two features sufficiently. Then energy based fusion is adopted to fuse features. The obtained features are concatenated with deconvolution. The final depth output is the sum of the learned residual depth and the coarse depth map.}
	\label{fig_pipeline}
\end{figure*}

\subsubsection{Feature Level Fusion} Approaches~\cite{yan2020revisiting}\cite{lee2020deep} for depth completion usually integrate depth and image information at the feature level. The image and depth features are extracted using two encoders, and a skip connection exists between the encoder and decoder parts. \cite{lee2020deep} proposes a cross-guidance module, and the image and depth features are fused through cross attention. Meanwhile,~\cite{yan2020revisiting} uses a Spatial Pyramid fusion (SPF) as a global attention block to merge the final outputs from two encoders.~\cite{van2019sparse}\cite{li2020multi}\cite{tang2019learning}\cite{tang2019learning} use image information to guide depth feature extraction for depth completion. Besides,~\cite{van2019sparse} uses global and local branches for depth completion, and the output of the image branch and the depth are connected as an input to the local branch. \cite{li2020multi} uses the cascade hourglass network to extract the multi-resolution depth map features for better depth completion. \cite{xu2019depth}\cite{qiu2019deeplidar} use surface normal to assist depth completion and \cite{chen2019learning} fuse the information between 2D and 3D spaces. However, all of the above methods ignore the integration of the color and depth information at the micro-level. Thus, the feature fusion is not sufficient, which limits the performance of these approaches. 

\subsubsection{Image Fusion}
Image fusion is the technique of integrating information on different types of images obtained from different sensors. \cite{liu2017multi}\cite{du2017image} propose a multi-focus image fusion method based on image segmentation through a multi-scale CNN. In \cite{prabhakar2017deepfuse}, the feature pairs of input images extracted from the last layer of the network are fused into a single feature by an addition operation. Instead of simply adding two features, \cite{li2018densefuse} apply a novel strategy based on $l1$-norm and soft-max operation into the network and get a better fusion result. \cite{liu2020wavefuse} proposes a multi-scene image fusion architecture based on the combination of the multi-scale discrete wavelet transform. The useful information of feature maps can be fully utilized, and a region-based fusion strategy is adopted to capture more detailed information.

However, the combination and fusion of color and depth information of the above approaches are insufficient, and to solve this, we propose a novel framework for better depth completion. 
\section{Our Approach}

We formulate depth completion as a two-stage task, including a sparse-to-coarse stage and a coarse-to-fine stage. The pipeline of our approach is demonstrated in Fig.~\ref{fig_pipeline}. First, a simple framework is utilized in the sparse-to-coarse stage to obtain coarse dense depth maps; Second, a coarse-to-fine stage is exploited, which contains a channel shuffle operation, an energy based fusion operation and a residual learning strategy. The channel shuffle extraction operation, which effectively extracts more representative features from color and depth information; The energy based fusion operation, which sufficiently fuses features obtained by channel shuffle extraction operation, thus better depth completion results can be expected; The residual learning strategy can further improve the quality of depth completion.

\subsection{Sparse-to-Coarse}

A dense depth map can be interpolated in handcrafted ways in the sparse-to-coarse stage, such as nearest-neighbor interpolation or other simple sparse-to-dense approaches. For the sparse-to-coarse stage, commonly used sparse-to-dense structures, such as~\cite{ma2019self}\cite{cheng2018depth}\cite{cheng2020cspn++}\cite{park2020non}, can be utilized, and to reduce the computational limitations, we use STD~\cite{ma2019self} in our approach. The process of the sparse-to-coarse stage can be formulated as:

\begin{equation}
    \begin{split}
        d_{sc} &= {SC}(d_s, I) 
    \end{split}
\end{equation}

where $d_s$ and $I$ mean the sparse depth and corresponding color image, ${SC}$ means the process of the sparse-to-coarse stage, and $d_{sc}$ means the obtained coarse dense depth map.

\subsection{Coarse-to-Fine}

The coarse-to-fine stage uses the color image and the corresponding coarse dense depth map as input, where the depth map is obtained after interpolation in the sparse-to-coarse stage. Thus consecutive information can be provided to the convolution. Meanwhile, to effectively and sufficiently extract and fuse features from color and depth information, a channel shuffle operation ($CS$) and an energy based fusion operation ($EF$) are utilized. Besides, a residual learning framework is exploited in the coarse-to-fine stage to improve the performance of depth completion further. In this section, we provide more details about these operations.

\subsubsection{Channel Shuffle}

Strategies~\cite{ma2019self}\cite{cheng2018depth}, have been proposed to extracted features from color and depth information with commonly used backbones, such as ResNet18 and ResNet34. These strategies usually stack color and depth directly and extract features using a single feature extractor, which performs the same feature extraction on information from different sources. Although information exchange exists in the process, some source-independent features cannot be extracted to a certain extent. Various strategies, such as~\cite{qiu2019deeplidar}\cite{lee2020deep}, extract features from color and depth information separately using two feature extractors, then fuse them with the same size using concatenate or add operation. However, the consistency of color and depth information is not utilized in the feature extraction process. Thus more representative features can not be obtained, which limits the performance of these approaches.

Inspired by~\cite{zhang2018shufflenet}, to well utilize the consistency of color and depth information, we propose to use channel shuffle extraction strategy in the coarse-to-fine stage, which extracts features from the color image and coarse dense depth map first, and then fully integrates the two different features at multi-scale channel levels.

Specifically, we use the commonly used backbones $R$, such as Resnet34~\cite{he2016deep}, to obtain the features with different sizes from color images and coarse depth maps, respectively. Given the input coarse depth map $d_{sc}$ and color image $I$, we define $f_d$ and $f_c$ as the extracted features, respectively. A convolution operation $Conv$ is first utilized to obtain features $f_{d_{0}} = Conv(d_{sc})$ and $f_{c_{0}} = Conv(I)$. We define the backbones used in color and depth feature extraction as $R_c = \{R_{c_{1}}, ..., R_{c_{N}}\}$, $R_d = \{R_{d_{1}}, ..., R_{d_{N}}\}$, where $N$ is the number of convolution blocks in the backbones. The features extracted by the first convolution block is defined as:
\begin{equation}
\begin{split}
    f_{d_{1}} &= R_{d_{1}}(f_{d_0}), \\
    f_{c_{1}} &= R_{c_{1}}(f_{c_0})
\end{split}
\end{equation}

The process of feature extraction by other convolution blocks can be formulated as:


\begin{equation}
\begin{split}
    (f^{'}_{d_{i-1}}, f^{'}_{c_{i-1}}) &= {CS}(f_{d_{i-1}}, f_{c_{i-1}}), \\
    f_{d_{i}} &= R_{d_{i}}(f^{'}_{d_{i-1}}), \\
    f_{c_{i}} &= R_{c_{i}}(f^{'}_{c_{i-1}}), \\
\end{split}
\end{equation}

where $i \in [2, N]$, and ${CS}$ means the channel shuffle operation, $N$ is the number of convolution blocks in the backbones.

\begin{figure}[tb!]  
	\centering
	\includegraphics[width=\linewidth]{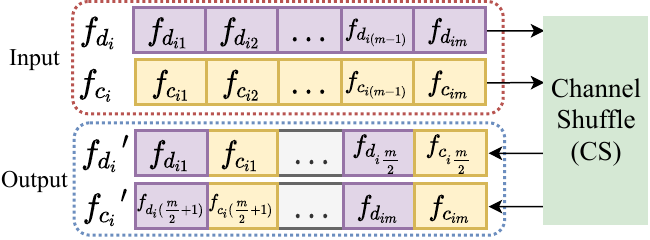}
	\caption{The proposed Channel shuffle operation. The purple $f_{d_i}$ and yellow $f_{c_i}$  denote depth and color image features, respectively. The channel shuffle (CS) obtains new features $f_{d_i}'$ and $f_{c_i}'$ by feature mixing in channel level, and return to their respective convolutions for the next step.}
	\label{channel_shuffle}
\end{figure}

The process of channel shuffle is shown in Fig.~\ref{channel_shuffle}, given depth and color features of the $i$-$th$ convolution block $f_{d_{i}} = \{f_{d_{i1}}, ..., f_{d_{iM}}\}$, $f_{c_{i}} = \{f_{c_{i1}}, ..., f_{c_{iM}}\}$, where $M$ is the number of channels, the output of channel shuffle are $f^{'}_{d_{i}} = \{f_{d_{i1}}, f_{c_{i1}}, ..., f_{d_{i\frac{M}{2}}}, f_{c_{i\frac{M}{2}}}\}$ and $f^{'}_{c_{i}} = \{f_{d_{i\frac{M}{2}+1}}, f_{c_{i\frac{M}{2}+1}}, ..., f_{d_{iM}}, f_{c_{iM}}\}$, respectively, which guarantees that features extracted by depth and color images are exchanged and mixed in channel level. Note that we assume $M$ is even number 
because the channel number is always even number in current DCNN based approaches. After mixing, two new feature maps $f^{'}_{di}$ and $f^{'}_{ci}$ are generated, and returned to the $(i+1)$-$th$ convolution block of backbones for next step. For each scale feature level, the different channels of the characteristics are fully mixed, which we call the channel shuffle operation. This operation can effectively extract new fusion features after shuffle and mixed. Experiments show that the result of this operation is significantly improved compared to the previous fusion methods.

\subsubsection{Energy based Fusion}

Inspired by~\cite{liu2020wavefuse}, features from large regional energy (pixel value) always contain more effective information. 
The residual learning strategy can be regarded as recovering high-frequency (HF) information of the depth map in the coarse-to-fine stage. Max pooling operation~\cite{boureau2010theoretical}\cite{max_pooling:striving} chooses large pixels and can well preserve texture information during the down-sampling process. Inspired by max pooling, to well recover HF and texture information, we propose a simple and effective energy based fusion operation to further sufficiently fuse the features $f_{d}$ and $f_{c}$ obtained by feature extraction.

Suppose that $H$, $W$ are the height and width of a feature map $f_{ij}$, where $i \in [0, N], j \in [1, M]$, and $N$ is the number of convolution blocks in backbone, $M$ is the number of channels, $f_{{k}_{ij}}(m,n)$ is the feature value at $(m,n)$, where $m \in [1, H]$, $n \in [1, W]$, and $f_k$ represents color and depth features, respectively. We define $E_k(m,n)$ to represent the energy in region ${L}\times{L}$ centered at $(m,n)$, and $k \in [1, 2]$ mean color and depth information, respectively. $E_k(m,n)$ can be computed as:


\begin{equation}
{E_{k_{ij}}(m,n)=\sum_{a=-m'}^{m'}\sum_{b=-n'}^{n'}\omega (f_{k_{ij}}(m+a,n+b))^2}
\label{eq:energy}
\end{equation}

where $m'=\lfloor \frac{L}{2} \rfloor$, $n'=\lfloor \frac{L}{2} \rfloor$, $\omega$ is a coefficient. The fusion feature map of color and depth can be represented as $f_o$, and $f_{o_{ij}}(m,n)$ ($i \in[0, N]$, $j \in[1, M]$) can be calculated as:


\begin{equation}
f_{o_{ij}}(m,n)=\left\{
             \begin{array}{lr}
             \sigma  f_{1_{ij}}(m,n),& E_{1_{ij}}(m,n)\geq E_{2_{ij}}(m,n) \\
             \sigma f_{2_{ij}}(m,n),& E_{1_{ij}}(m,n)\textless E_{2_{ij}}(m,n)\\
             \end{array}
\right.
\label{eq:fusion_choose}
\end{equation}

where $\sigma$ is a coefficient. In this paper, we set $L=5$, $\omega=1$, $\sigma=2$, empirically, and we will provide more analysis in the supplementary material. According to Eq.~\ref{eq:energy} and Eq.~\ref{eq:fusion_choose}, the features extracted from color and depth information can be sufficiently fused. By selecting the feature value with higher regional energy instead of add or concatenate the two features, the feature fusion result can be effectively improved and get more useful information.

\subsubsection{Residual Learning}

In the training process, given the features $f$ obtained by energy based fusion operation, a convolution operation $Conv$ is used to obtain the residual depth $d_r$. And the final output $d_o$ of the coarse-to-fine stage can be obtained by:


\begin{equation}
\begin{split}
    d_r &= Conv(f), \\
    d_o &= d_r + d_{sc}
\end{split}
\end{equation}

\subsection{Loss Function}

The loss used in the coarse-to-fine stage is defined as:

\begin{equation}
Loss = \frac{1}{\nu} \sum^{H_d}_m \sum^{W_d}_n {\left|(d_{o}(m,n)-d_{gt}(m,n)) \times \Psi \right|^p}
\label{eq:loss}
\end{equation}

where $d_o$ and $d_{gt}$ mean the recovered depth map and ground truth, respectively, $H_d$ and $W_d$ are the height and width of the depth map, and $p\in{\{1,2\}}$, if the value of the corresponding position of ground truth is valid, the value of the position of $\Psi$ is 1, and the rest is zero. $\nu$ denotes the number of non-zero points in the $\Psi$.

\section{Experiment}

We evaluate the performance of our method against different state-of-the-art (SoTA) methods on diverse publicly available datasets, including the KITTI and NYUDv2 dataset.

\subsection{KITTI Dataset and Implementation Details}

The KITTI dataset~\cite{Geiger2013IJRR} is a large outdoor dataset for autonomous driving, which contains 85k color images and corresponding sparse depth maps for training, 6k for validation, and 1k for testing. In validation, 1000 color images and corresponding sparse depth maps are selected as validation data. For training, we bottom-cropped color and depth images to $352\times1216$.

In the sparse-to-coarse stage, the framework of STD~\cite{ma2019self} is used as a simple network which pre-trained on KITTI to obtain a coarse dense depth map, and other approaches can also be used here. All models are trained with Adam optimizer with $\beta_1$=0.9, $\beta_2$=0.999. We set batch size as 8, the learning rate starts from 1e-5 and reduces by 0.1 for every 10 epochs. The $p$ in the loss function is set to 2. The models are trained for 20 epochs.

\subsection{NYUDv2 Dataset and Implementation Details}
The NYUDv2\cite{silberman2012indoor} dataset is comprised of video sequences from a variety of indoor scenes as recorded by both the color and depth cameras from the Microsoft Kinect. Following~\cite{mal2018sparse}\cite{cheng2018depth}\cite{cheng2020cspn++}, we utilize a subset of 45K images from the official training split as training data, and 654 official labeled images are used for evaluation. Since the input resolution of our network must be a multiple of 16, for a fair comparison with other methods, we first down-sampled the input frames to $320\times240$, and center-cropped the prediction of the network to $304\times228$ during evaluation.

Like the KITTI dataset, the framework of STD~\cite{ma2019self} is used as the simple network to obtain coarse dense depth maps. The $p$ in the loss function is set to 1. The learning rate starts from 1e-5 and reduces by 0.1 for every 10 epochs, and the model is trained for 20 epochs. We utilize the Adam as the optimizer with $\beta_1$=0.9, $\beta_2$=0.999, weight-decay=0.01. When training the fine network, we freeze the parameters of the coarse network and finally make end-to-end predictions during the evaluation.

\subsection{Evaluation Metrics}
We use the standard metrics for evaluation: (1) root mean squared error (RMSE): $ \sqrt{\frac{1}{\nu} \sum_x (\hat{d}_x-d_x)^2}$; (2) mean absolute error (MAE): $\frac{1}{\nu} \sum_x \left|\hat{d}_x-d_x\right|$; (3) root mean squared error of the inverse depth (iRMSE): $\sqrt{\frac{1}{\nu} \sum_x (\frac{1}{\hat{d}_x}-\frac{1}{d_x})^2}$; (4) mean absolute error of the inverse depth (iMAE) $\frac{1}{\nu} \sum_x \left|\frac{1}{\hat{d}_x}-\frac{1}{d_x}\right|$.

For NYUDv2, in addition to using RMSE as an evaluation metric, there are also the following: (1) mean absolute relative error (REL): $ \frac{1}{\nu} \sum_x \left|\frac{\hat{d}_x-d_x}{d_x}\right|$; (2) $\delta_\tau$: Percentage of pixels satisfying $max(\frac{d_x}{\hat{d}_x},\frac{\hat{d}_x}{d_x})<\tau$, $\tau\in{\{1.25, 1.25^2, 1.25^3\}}$.

\subsection{Evaluation on KITTI Dataset}

\begin{figure*}[thb!]
	\centering
	\setlength{\tabcolsep}{1pt}
	\def\myheight{.077} 
	\begin{tabular}{cccccc} 
		{(a)} &
		\includegraphics[height=\myheight\linewidth]{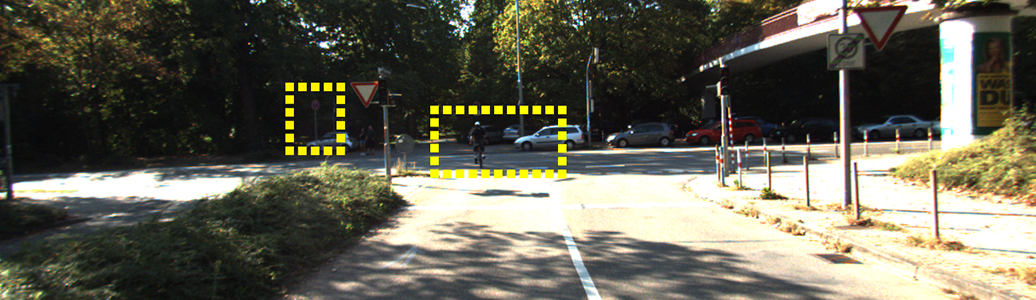} &
		\includegraphics[height=\myheight\linewidth]{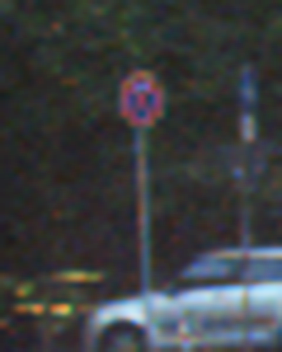} &
		\includegraphics[height=\myheight\linewidth]{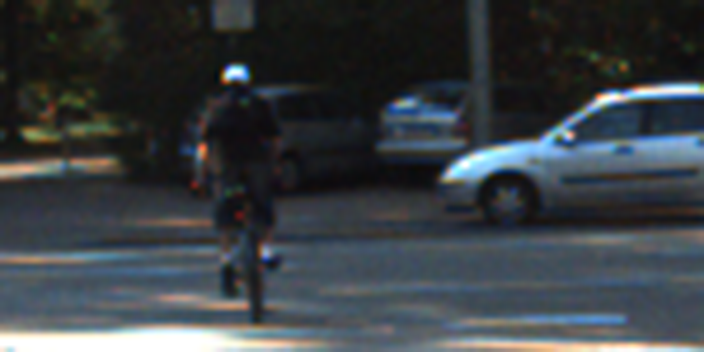} &
		\includegraphics[height=\myheight\linewidth]{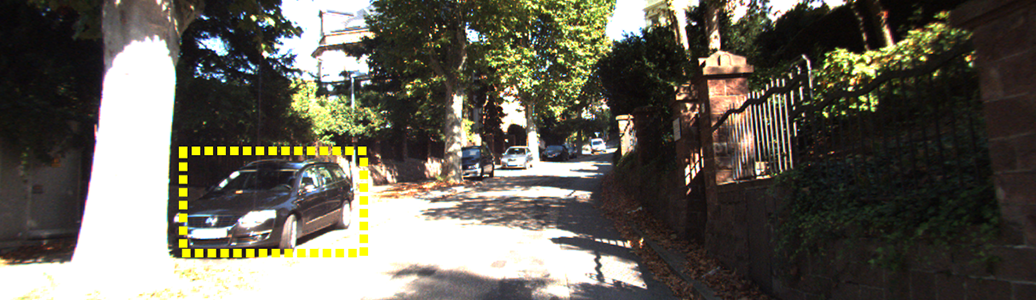} &
		\includegraphics[height=\myheight\linewidth]{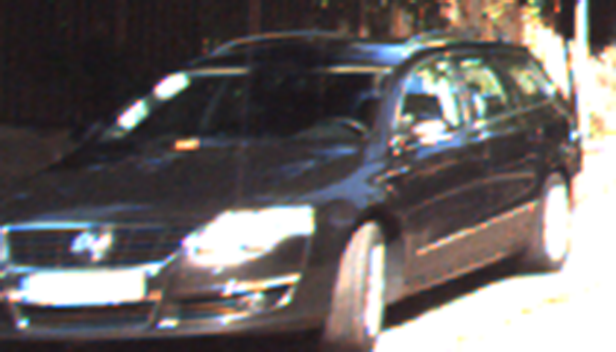} \\
		{(b)} &
		\includegraphics[height=\myheight\linewidth]{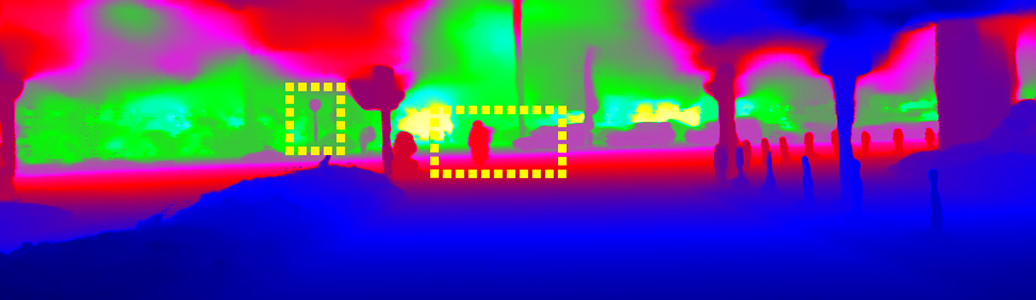} &
		\includegraphics[height=\myheight\linewidth]{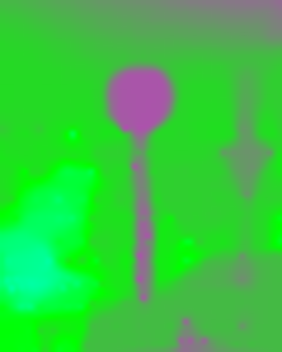} &
		\includegraphics[height=\myheight\linewidth]{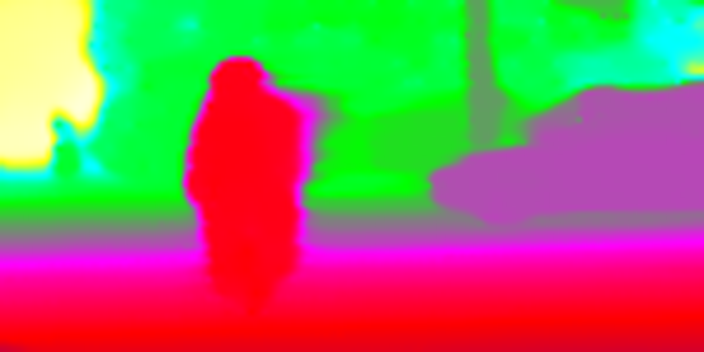} &
		\includegraphics[height=\myheight\linewidth]{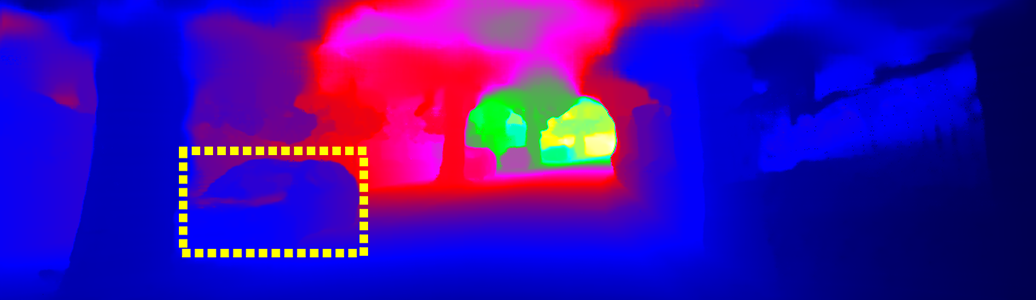} &
		\includegraphics[height=\myheight\linewidth]{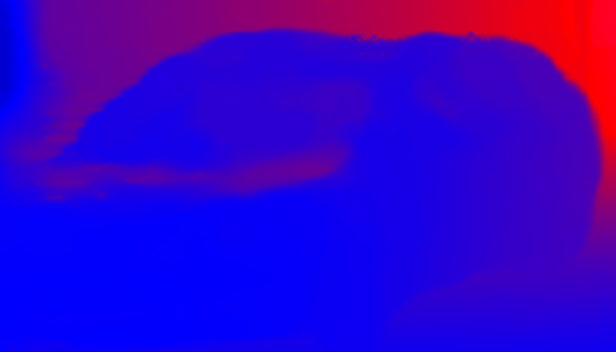} \\
		{(c)} &
		\includegraphics[height=\myheight\linewidth]{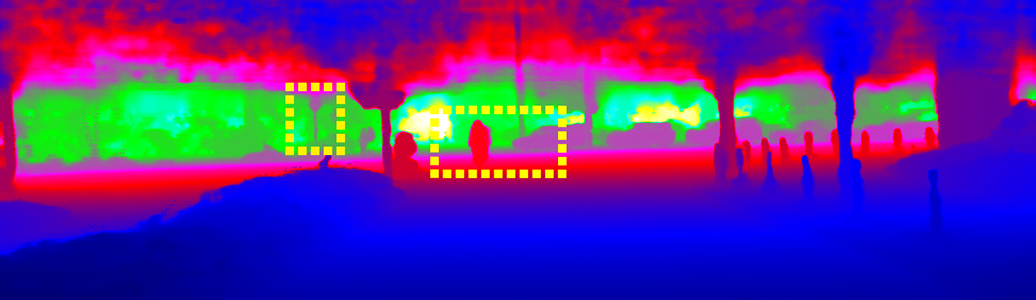} &
		\includegraphics[height=\myheight\linewidth]{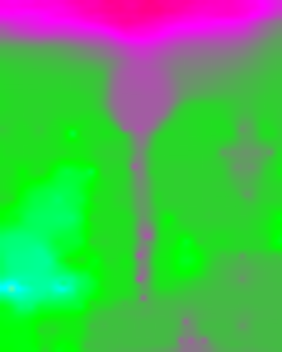} &
		\includegraphics[height=\myheight\linewidth]{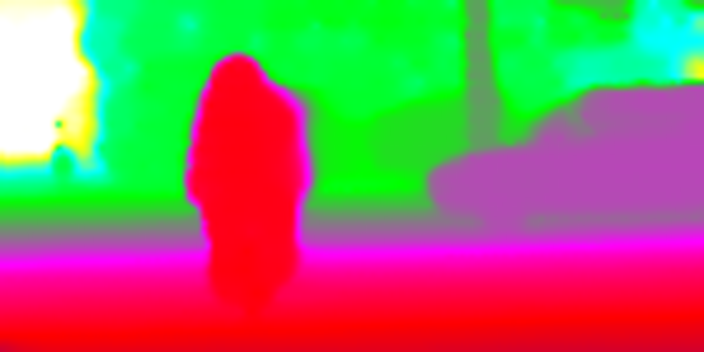} &
		\includegraphics[height=\myheight\linewidth]{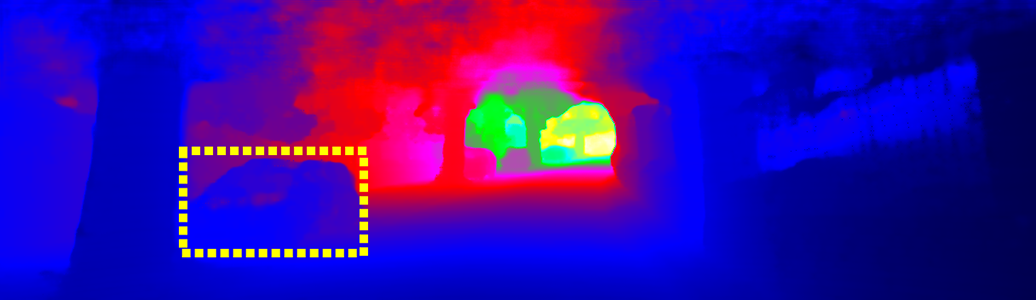} &
		\includegraphics[height=\myheight\linewidth]{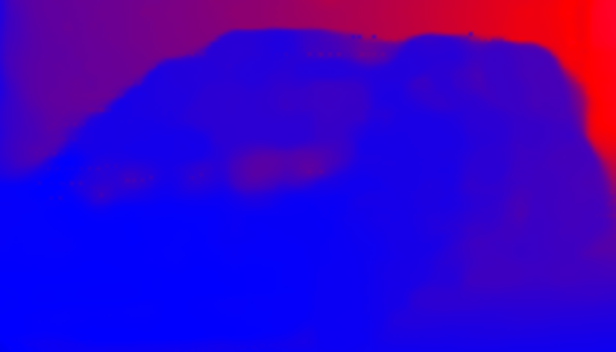} \\
		{(d)} &
		\includegraphics[height=\myheight\linewidth]{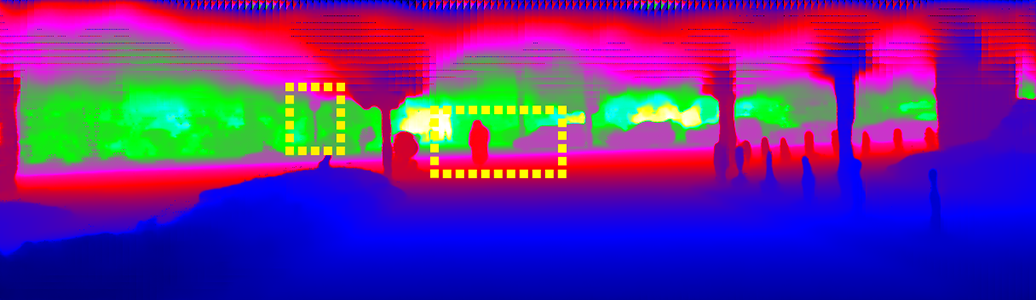} &
		\includegraphics[height=\myheight\linewidth]{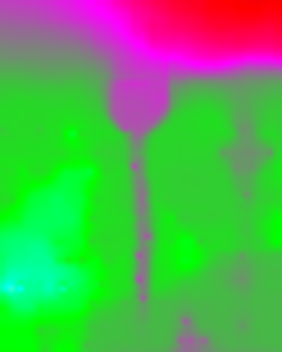} &
		\includegraphics[height=\myheight\linewidth]{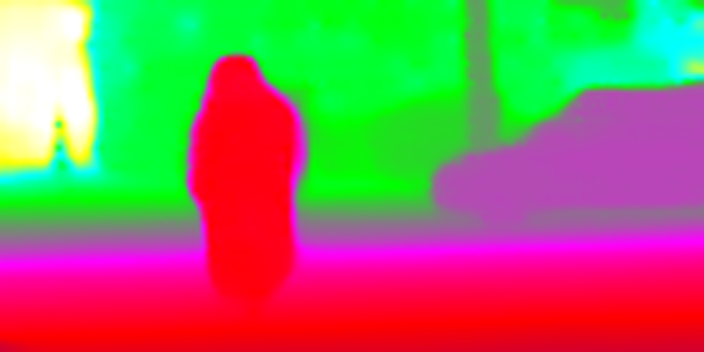} &
		\includegraphics[height=\myheight\linewidth]{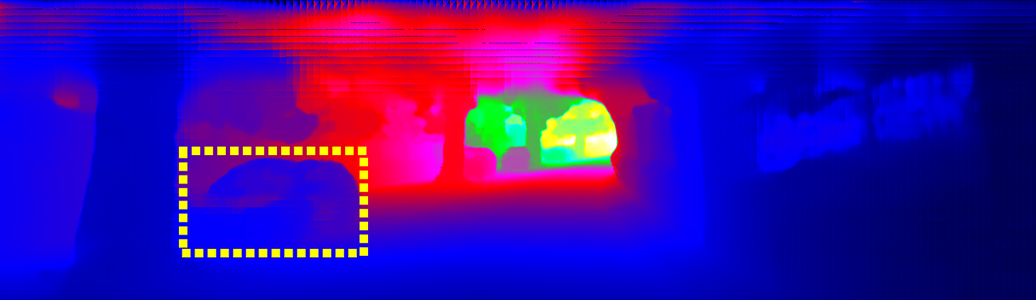} &
		\includegraphics[height=\myheight\linewidth]{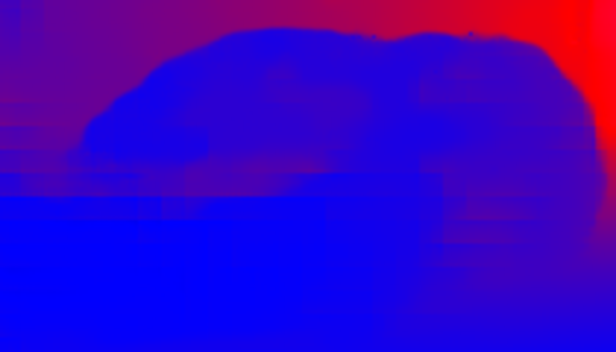} \\
		{(e)} &
		\includegraphics[height=\myheight\linewidth]{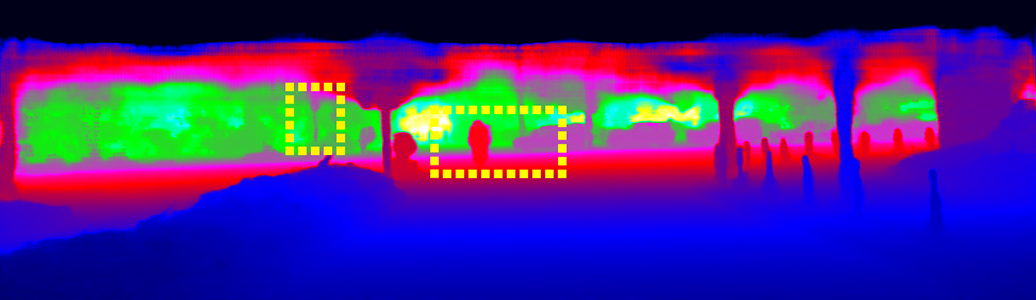} &
		\includegraphics[height=\myheight\linewidth]{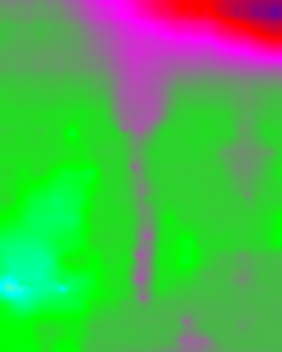} &
		\includegraphics[height=\myheight\linewidth]{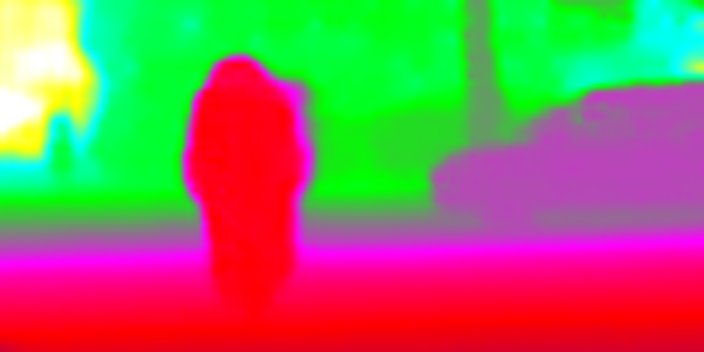} &
		\includegraphics[height=\myheight\linewidth]{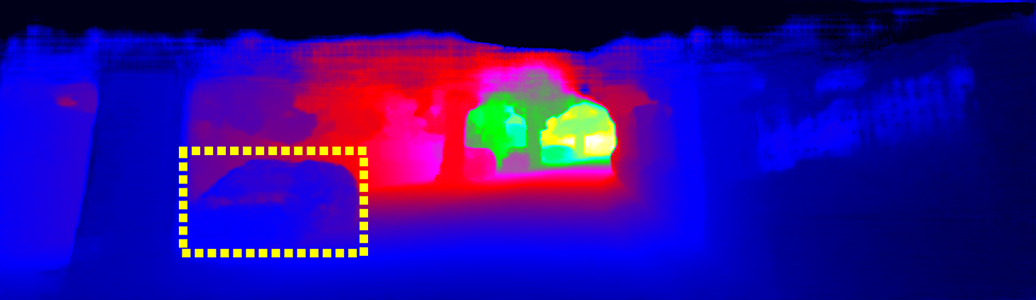} &
		\includegraphics[height=\myheight\linewidth]{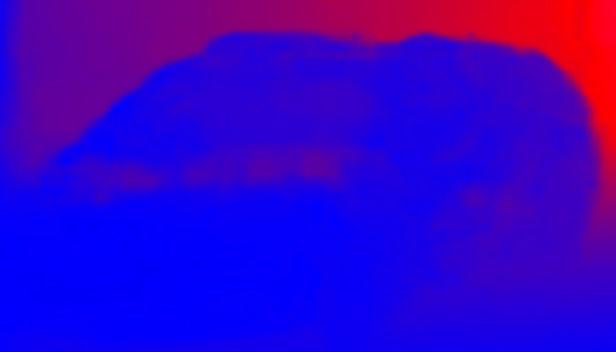} \\
		{(f)} &
		\includegraphics[height=\myheight\linewidth]{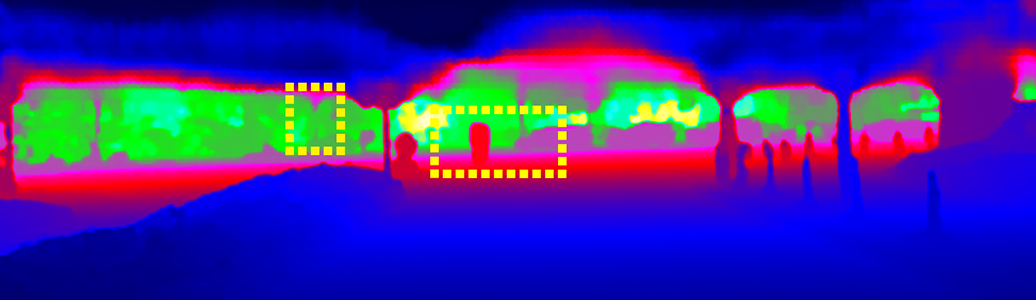} &
		\includegraphics[height=\myheight\linewidth]{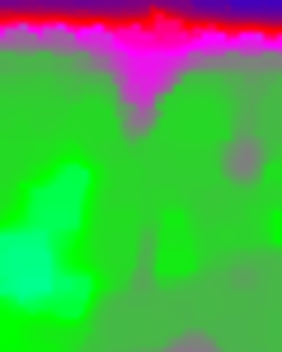} &
		\includegraphics[height=\myheight\linewidth]{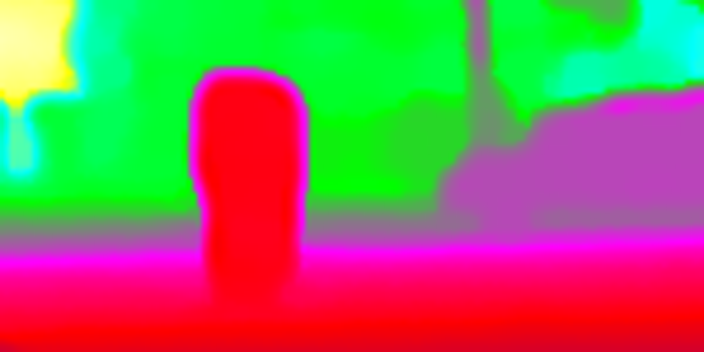} &
		\includegraphics[height=\myheight\linewidth]{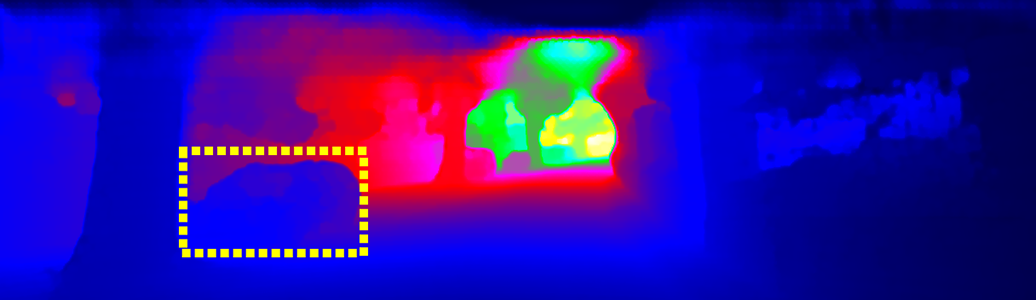} &
		\includegraphics[height=\myheight\linewidth]{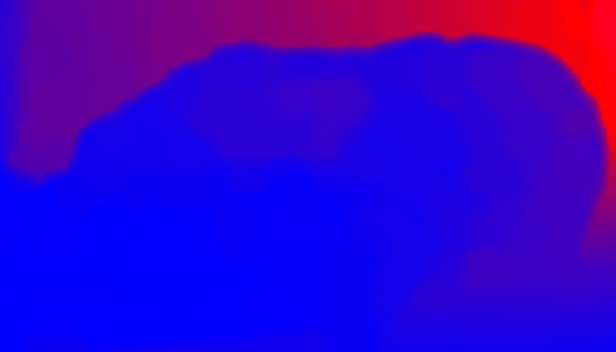} \\
		{(g)} &
		\includegraphics[height=\myheight\linewidth]{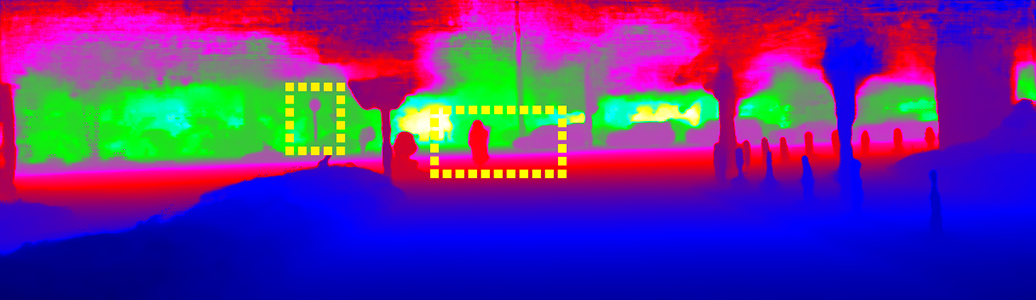} &
		\includegraphics[height=\myheight\linewidth]{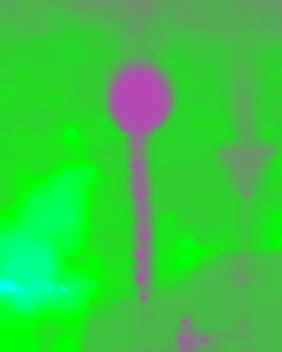} &
		\includegraphics[height=\myheight\linewidth]{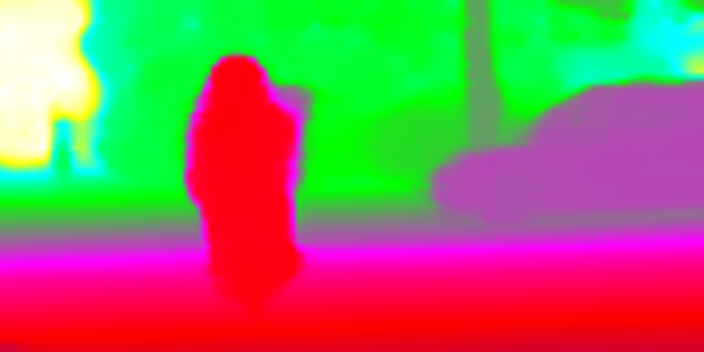} &
		\includegraphics[height=\myheight\linewidth]{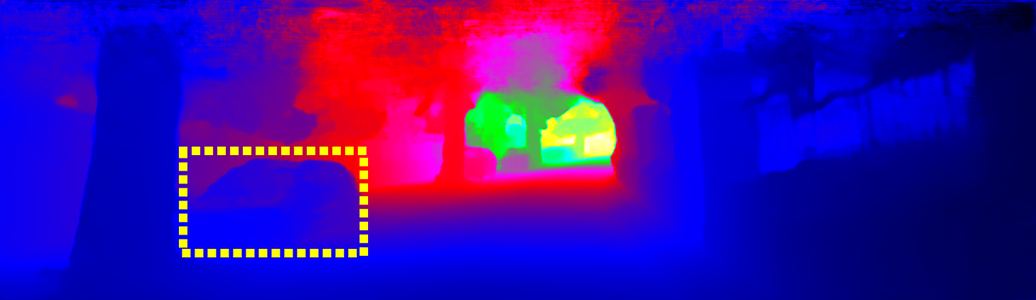} &
		\includegraphics[height=\myheight\linewidth]{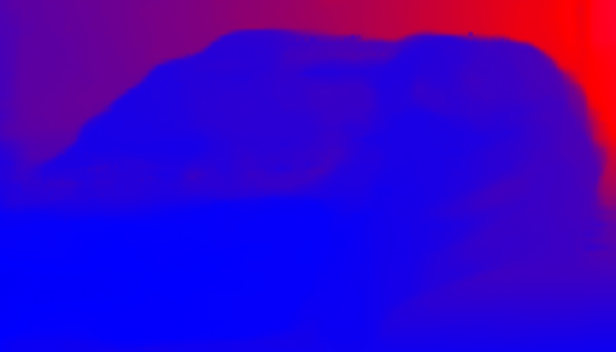} \\
		
	\end{tabular}
	
	
	\caption{Qualitative Comparison on KITTI test set. From top-to-bottom: (a) Image, (b) NLSPN \cite{park2020non}, (c) CSPN++ \cite{cheng2020cspn++}, (d) DeepLidar \cite{qiu2019deeplidar}, (e) STD \cite{ma2019self}, (f) CSPN \cite{cheng2018depth}, (g) Ours (FCFR-Net). The results are from the KITTI depth completion leaderboard in which depth images are colorized along with depth range. }

	\label{fig:kitti_evaluation}
\end{figure*}

Table.~\ref{table:kitti_evaluation} demonstrates the quantitative comparison results of our approach on the KITTI benchmark. Note that the results of STD is obtained in supervised manner in Table~\ref{table:kitti_evaluation} and Table.~\ref{table:nyudv2_evaluation}. It is obvious to find that our FCFR-Net outperforms existing SoTA approaches in RMSE, which is the main evaluation metric on the KITTI depth completion benchmark. Due to the sensitivity of RMSE to outliers, our approach has better processing ability for long-distance depth. Compared with the results of STD~\cite{ma2019self}, our coarse-to-fine stage improves performance by about 10\% in RMSE. The qualitative comparison is shown in Fig.~\ref{fig:kitti_evaluation}. We can find that depth maps obtained by our approach are with sharper boundaries and more depth details, especially on long-distance, which proves the effectiveness of our approach.

\begin{table}[htb!]
    \small
	\begin{center}

\begin{tabular}{p{3.7cm} |  p{0.7cm}<{\centering} p{0.7cm}<{\centering} p{0.6cm}<{\centering} p{0.6cm}<{\centering}}
	
	\hline
    \multirow{2}{*}{Method}   &  RMSE &         MAE & iRMSE     &     iMAE      \\
	                        &    mm  &          mm&  1/km    &     1/km       \\ 
	\hline
	CSPN 	& 1019.64 &  279.46 & 2.93 & 1.15  \\ 
	STD  &  814.73&  249.95 & 2.80 & 1.21  \\
	CG~\cite{lee2020deep}		& 807.42 &		253.98&	2.73&	1.33 \\ 
	RV		& 792.80 &		225.81&	2.42&	0.99 \\ 
	PwP	\cite{xu2019depth}		& 777.05 &	235.17&	 2.42	&1.13 \\
	RGBG\&C		&772.87&		215.02 &	2.19&	0.93 \\
    MSG-CHN~\cite{li2020multi}		&762.19&	220.41	&	2.30&	0.98	\\
	DeepLiDAR~\cite{qiu2019deeplidar}		&758.38&		226.50&	2.56&	1.15 \\
    Uber~\cite{chen2019learning}		& 752.88&		221.19&	2.34&	1.14 \\
    CSPN++~\cite{cheng2020cspn++}			& 743.69&209.28	& 2.07	&0.90	 \\
    NLSPN~\cite{park2020non}              & 741.68 &  \textbf{199.59}& \textbf{1.99} & \textbf{0.84}   \\
    Ours                                & \textbf{735.81} & 	217.15&  2.20	& 	0.98 \\
	\hline

\end{tabular}

	

	\end{center}
	\caption{Quantitative comparison with state-of-the-art methods on KITTI Depth Completion testing set. The results of other methods are obtained from the KITTI online evaluation site. The CSPN, STD, RV and RGBG\&C mean~\cite{cheng2018depth}\cite{ma2019self}\cite{yan2020revisiting}\cite{van2019sparse}, respectively. The results are ranked by the RMSE.}
	\label{table:kitti_evaluation}
\end{table}

\subsection{Evaluation on NYUDv2 Dataset}

\begin{figure*}[thb!]
	\centering
	\setlength{\tabcolsep}{1pt}
	\def\mywidth{.132} 
	\begin{tabular}{ccccccc} 

        \includegraphics[width=\mywidth\linewidth]{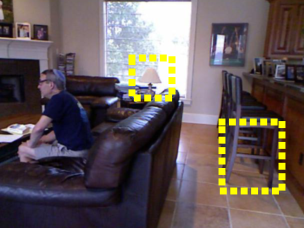} &
		\includegraphics[width=\mywidth\linewidth]{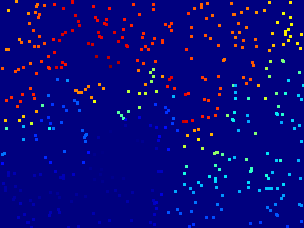} &
		\includegraphics[width=\mywidth\linewidth]{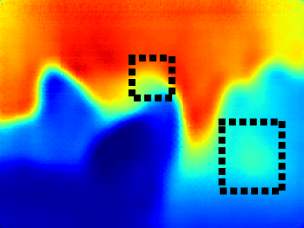} &
		\includegraphics[width=\mywidth\linewidth]{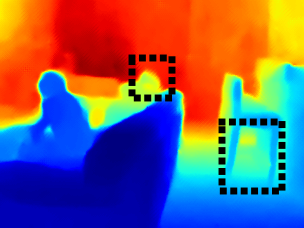} &
		\includegraphics[width=\mywidth\linewidth]{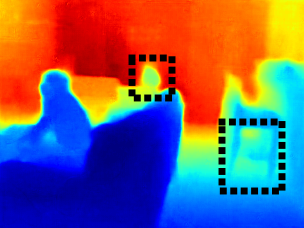} &
		\includegraphics[width=\mywidth\linewidth]{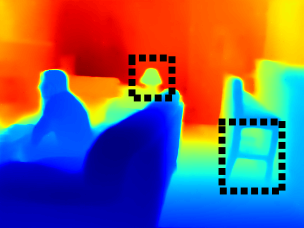} &
		\includegraphics[width=\mywidth\linewidth]{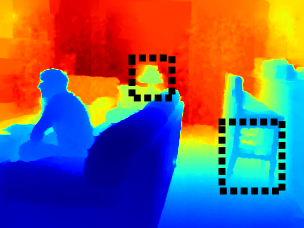} \\
		
		\includegraphics[width=\mywidth\linewidth]{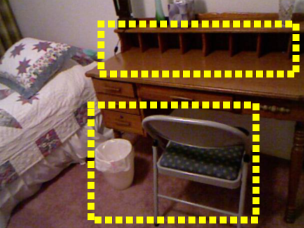} &
		\includegraphics[width=\mywidth\linewidth]{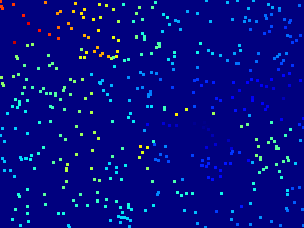} &
		\includegraphics[width=\mywidth\linewidth]{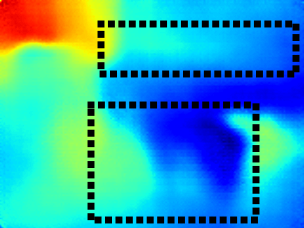} &
		\includegraphics[width=\mywidth\linewidth]{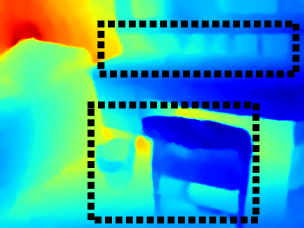} &
		\includegraphics[width=\mywidth\linewidth]{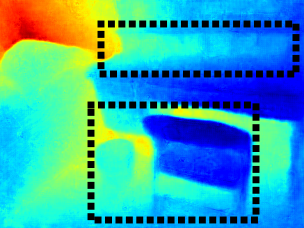} &
		\includegraphics[width=\mywidth\linewidth]{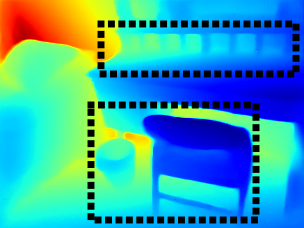} &
		\includegraphics[width=\mywidth\linewidth]{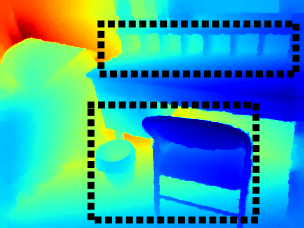} \\
		(a) & (b) & (c) & (d) & (e) & (f) & (g) \\
        
	\end{tabular}
	\caption{Qualitative Comparison on NYUDv2. From left-to-right: (a) Image, (b) Dilated sparse input for visualization, (c) Sparse-to-dense \cite{mal2018sparse}, (d) CSPN \cite{cheng2018depth}, (e) Coarse result using Sparse-to-dense(gd) \cite{ma2019self}, (f) Ours fine result, (g) Ground Truth. The circled  rectangles areas show the recovery of object details. }

	\label{fig:nyud_evaluation}
\end{figure*}

To verify the effectiveness of our approach, we also evaluate our approach on the NYUDv2 dataset. Following CSPN++~\cite{cheng2020cspn++} and NLSPN~\cite{park2020non}, we use 500 randomly sampled points as sparse input and the quantitative comparisons results are shown in Table.~\ref{table:nyudv2_evaluation}. It can be seen that our approach outperforms SoTA approaches in all metrics with 500 sampling points. And compared with STD~\cite{ma2019self}, the RMSE results obtained by our approach decrease by 13\%. Fig.~\ref{fig:nyud_evaluation} illustrates the qualitative comparison results, and depth maps obtained by our approach have more depth details and object edge structures, which proves the effectiveness of our approach on indoor scenes. 


\begin{table}[htb!]
    \small
	\begin{center}
		
\begin{tabular}{p{3.7cm} | p{0.5cm}<{\centering} p{0.5cm}<{\centering} p{0.4cm}<{\centering} p{0.4cm}<{\centering} p{0.4cm}<{\centering}}
	
	\hline
    \multirow{2}{*}{Method}   &   RMSE     &     REL   & $\delta_{1.25}$&$\delta_{1.25^2}$&$\delta_{1.25^3}$   \\
	                        &     m    &     m   &      &         &   \\ 
	\hline
	STD\_18                      & 0.230 & 0.044& 97.1 & 99.4 & 99.8        \\
	Sparse-to-Coarse   & 0.123 & 0.026 & 99.1 & \textbf{99.9} & \textbf{100.0} \\
	CSPN 	&0.117 & 0.016 & 99.2 & \textbf{99.9} & \textbf{100.0} \\ 
	CSPN++~\cite{cheng2020cspn++} 	            &0.116 & - & - & - & - \\
	DeepLiDAR~\cite{qiu2019deeplidar} 	            &0.115 & 0.022 & 99.3 & \textbf{99.9} & \textbf{100.0} \\
	PwP~\cite{xu2019depth} 	            &0.112 & 0.018 & \textbf{99.5} & \textbf{99.9} & \textbf{100.0} \\


	Ours                     &\textbf{0.106}	& \textbf{0.015}&  \textbf{99.5} & \textbf{99.9} & \textbf{100.0}\\
	\hline

\end{tabular}
	\end{center}
	\caption{Quantitative evaluation on the NYUDv2 dataset. Sparse-to-Coarse is trained using STD \cite{ma2019self}, other methods are the results of the respective papers. STD\_18 means~\cite{mal2018sparse}, CSPN means~\cite{cheng2018depth}. And all methods use 500 sampled depth points as the sparse input. }
	\label{table:nyudv2_evaluation}
\end{table}

\begin{table*}[htb!]
    \small
	\begin{center}
		
\begin{tabular}{ c c c c c c |c c c c}
	
	\hline
     \multirow{2}{*}{name} & \multirow{2}{*}{S/D input} &\multirow{2}{*}{residual} &\multirow{2}{*}{number of encoders} &\multirow{2}{*}{CS} & \multirow{2}{*}{EF} &   iRMSE     &     iMAE   &   RMSE &      MAE   \\
	                        & &&&& &   1/km    &     1/km   &     mm &      mm    \\ 
	\hline
	$SI$ & S &  N& 1& N& N (concat)&             3.81 & 1.60 & 942.65 & 299.46\\
	$DI$ & D &  N& 1& N& N (concat)&             2.90 & 1.35 & 845.55 & 264.80 \\
	$DR$ & D &  Y& 1& N& N (concat)&             2.48 &  1.03 & 816.94 & 228.27 \\
	$DE$ & D &  Y& 2& N& N (concat)&             2.41 & 1.00 & 812.92 & 224.80 \\

	$DCC$ & D &  Y & 2 & Y &  N (concat) & 2.41 & 1.00 & 805.23 & 224.07  \\
	$DCA$ & D &  Y & 2 & Y &  N (add) &  2.40 & 1.01 & 806.26 & 225.23  \\

	$DCE$ & D &  Y & 2 & Y & Y  & 2.39 & 1.00 & 802.62 & 224.53 \\

	\hline

\end{tabular}
	\end{center}
	\caption{Ablation study on KITTI Depth Completion selected validation dataset. $CS$ means channel shuffle, and $EF$ means energy based fusion.}
	\label{table:ablation_study}
\end{table*}

\subsection{Ablation Studies}

In this section, we provide more analyses of the operations utilized in our approach. We sort the depth maps in time series and uniformly sample $1/4$ of the data as mini-training data for ablation studies for fast training.

\subsubsection{Sparse Input vs. Dense Input}

Using STD~\cite{ma2019self} as the baseline, we compare the depth completion results between sparse depth and dense depth as input. The output of STD~\cite{ma2019self} (with sparse depth as input) is regarded as the dense depth map. Under the same framework, the results are shown in Table.~\ref{table:ablation_study} ($SI$ and $DI$), where $S$ stands for sparse depth input, and $D$ stands for dense depth input. We can see that when the input depth is dense, all evaluation metrics are greatly reduced, which indicates that the dense depth input provides effective consecutive information. Thus better depth completion results can be obtained.

\subsubsection{Directly Learning vs. Residual Learning}

When the input depth map is dense, network learning has two choices: directly learn the final depth, or learn the residual between the input depth map and the ground truth. In Table.~\ref{table:ablation_study} ($DI$ and $DR$), we compare the results between direct learning and residual learning. Compared with direct learning, it is easy to find that residual learning results are greatly improved in all evaluation metrics.

\subsubsection{One Feature Extractor vs. Two Feature Extractors}

For feature extraction, there are two choices: (1) concatenate color and depth images and use a feature extractor to extract features, which named one feature extractor; (2) use two feature extractors to extract features of color and depth information respectively, then fuse them with concatenating operation, which named two feature extractors. Table.~\ref{table:ablation_study} $DR$ and $DE$ show the results of one feature extractor and two feature extractors, where the two encoders can extract features of different scales of color and depth, and merge them separately, and the result is better than one encoder.

\subsubsection{Channel Shuffle vs. No Channel Shuffle}

To demonstrate the effectiveness of the proposed channel shuffle operation, the depth completion results with and without channel shuffle are shown in Table.~\ref{table:ablation_study} ($DCC$ and $DE$). We can see that results obtained by approaches with channel shuffle outperform no channel shuffle, which proves that the channel shuffle operation can sufficiently exchange and fuse the features of color and depth information. Thus more representative features and better depth completion results can be obtained.

\subsubsection{Energy based Fusion vs. No Energy based Fusion}

DCNN based approaches usually fuse the features extracted from color and depth information with concatenating or add operation. Here, to prove the effectiveness of the proposed energy based fusion operation, we compare the results obtained by energy based fusion with other fusion operations (concatenate and add) in Table.~\ref{table:ablation_study}. $DCC$ means concatenate fusion results, $DCA$ means add fusion results, and $DCE$ means energy based fusion results. It can be seen that energy based fusion achieves better results for all evaluation metrics, which proves that the proposed energy based fusion operation can sufficiently fuse the features extracted from color and depth information, thus obtain better depth results.

\section{Conclusion}
In this paper, we propose a simple and effective framework for depth completion, which tackles the problem as a two-stage task, i.e., a sparse-to-coarse stage and a coarse-to-fine stage. We find that dense depth maps can provide consecutive features; thus, better depth results can be obtained. Meanwhile, to obtain more representative features, channel shuffle and energy based fusion operations are proposed, which effectively and sufficiently extract and fuse the features with color and depth images as input. Thus more accurate depth completion results can be achieved. Extensive experiments across indoor and outdoor benchmarks demonstrate the superiority of our approach over state-of-the-art approaches.

\section{Acknowledgements}

This work is supported by Robotics and Autonomous Driving Lab of Baidu Research. Besides, the work is also supported in part by the Key Research and Development Program of Guangdong Province of China (2019B010120001) and the National Natural Science Foundation of China under Grant 61836015.

\bibliography{bibliography}

\end{document}